\documentclass[11pt]{article}

\usepackage[final]{acl}

\usepackage{fontspec}
\usepackage{polyglossia}

\defaultfontfeatures{Ligatures=TeX}

\setmainlanguage{english}
\setotherlanguage{arabic}
\newfontfamily\arabicfont[
  Script         = Arabic,
  BoldFont       = Amiri-Bold.ttf,
]{Amiri-Regular.ttf}

\usepackage{latexsym}
\usepackage{amsmath}
\usepackage{amssymb}
\usepackage{subcaption}
\usepackage{booktabs}
\usepackage{multirow}
\usepackage{graphicx}   
\usepackage{colortbl}
\usepackage{dblfloatfix}
\usepackage{float}
\usepackage[table]{xcolor}
\definecolor{shaderow}{gray}{0.93}

\usepackage{microtype}

\usepackage{inconsolata}

\usepackage{listings, xcolor}

\title{Noise Steering for Controlled Text Generation: Improving Diversity and Reading-Level Fidelity in Arabic Educational Story Generation}

\author{\AND Haziq Mohammad Khalid \\
  \texttt{b00100601@aus.edu} \\\And
  Salsabeel Shapsough \\
  \texttt{sshapsough@aus.edu} \\\And
  {\bf Imran Zualkernan} \\
  \texttt{izualkernan@aus.edu} \\ \AND
  \\ {American University of Sharjah}
  \AND
  }

\begin{document}
\maketitle

\begin{abstract}
    Generating diverse, pedagogically valid stories for Arabic early-grade reading assessments requires balancing tight constraints on vocabulary, reading level, and narrative structure against the need to avoid repetitive plots that undermine assessment validity. We investigate \textit{noise steering}, injecting calibrated Gaussian perturbations into the internal representations of transformer models at inference time, as a training-free diversity method evaluated across five small Arabic-centric language models (7–9B parameters). We compare four injection strategies against high-temperature sampling baselines, measuring diversity, quality, constraint adherence, and reading grade level. Residual stream noise consistently improves narrative diversity with minimal quality or constraint cost and preserves early-grade reading level across all Arabic-centric models. Attention entropy noise injection (AENI) stabilizes the otherwise unreliable attention-logit noise while recovering quality. High-temperature sampling inflates reading grade level and causes catastrophic collapse on several models. We find internal representation-level perturbation to be a more suitable diversity strategy than output-level stochasticity for constrained educational content generation.
\end{abstract}

\section{Introduction}

The Early Grade Reading Assessment (EGRA) is a widely used framework for evaluating foundational reading skills in young learners \cite{dubeck2015egra}. A key component involves short narrative passages that must satisfy strict pedagogical constraints: controlled vocabulary, appropriate reading level, clear narrative structure, and cultural neutrality. Manually producing such passages at scale is labor-intensive, making automatic generation an attractive alternative, particularly in Arabic, where rich morphology, orthographic variation, and diglossia introduce additional linguistic complexity \cite{habash2010introduction}.

Recent work has shown that large language models can generate fluent Arabic stories aligned with EGRA criteria, but two tensions persist \cite{abdelghafur2025tales,rai2024measuring}. First, \textit{diversity vs. constraint adherence}: increasing output variety risks violating the tight pedagogical requirements that make EGRA passages valid assessment instruments \cite{holtzman2020curious}. Second, practical deployment in under-resourced educational settings favors small language models (SLMs) of 7–9 billion parameters over large frontier models, yet these smaller Arabic-centric models are more susceptible to generation instability when diversity is pushed.
We address both tensions by investigating \textit{noise steering}: injecting calibrated Gaussian perturbations into the internal representations of transformer models at inference time, without any fine-tuning. We evaluate four injection strategies across five Arabic SLMs and show that residual stream noise and attention entropy noise injection (AENI) consistently improve narrative diversity while preserving reading grade level and constraint adherence, outperforming output-level approaches such as high-temperature sampling that raise diversity at the cost of pedagogical validity.

\section{Background}

\subsection{EGRA Constraints}

EGRA-aligned texts are constructed under strict pedagogical and linguistic constraints \cite{dubeck2015egra}. These include controlled passage length, age-appropriate vocabulary, a clear narrative structure, limited characters, grammatical correctness, and cultural neutrality. These constraints ensure comparability across assessments and support the generation of both literal and inferential comprehension questions. However, manually creating such passages is labor-intensive and requires domain expertise, limiting scalability \cite{abdelghafur2025tales}.

\subsection{Challenges in Arabic EGRA Generation}
Generating EGRA-aligned content in Arabic introduces additional challenges due to the language's rich morphology, orthographic ambiguity, and diglossia between Modern Standard Arabic and dialectal varieties, compounded by the limited availability of high-quality annotated datasets for children's literature \cite{habash2010introduction,abdelghafur2025tales}. While recent work has shown that models such as GPT-4 and Jais can generate fluent, EGRA-aligned stories, inconsistencies in narrative structure, deviations from target word counts, and repetitive outputs with minimal structural variation remain persistent issues \cite{abdelghafur2025tales,rai2024measuring,elshangiti2024arabic}.





\subsection{Diversity vs. Constraint Trade-off}

A central challenge in EGRA-based generation is balancing diversity with constraint satisfaction. Increasing diversity through stochastic decoding can lead to output degeneration and reduced control over pedagogical requirements \cite{holtzman2020curious}, while overly constrained prompting often results in repetitive and structurally similar outputs \cite{rai2024measuring}. 

Recent approaches attempt to address this trade-off. \citet{shankarnarayanan2024once} show that inspiration-based prompting can improve narrative diversity, while \citet{madaan2023selfrefine} demonstrate that iterative self-refinement can enhance output quality without model fine-tuning. Despite these advances, achieving consistent diversity while maintaining strict adherence to EGRA constraints remains unresolved, particularly in Arabic.

\section{Related Works}

\subsection{Output-Level Diversity: Stochastic Decoding Strategies}

A common approach to diversity in LLM generation is stochastic decoding, which modifies the token probability distribution at each step. Temperature scaling \cite{ackley1985learning} flattens or sharpens this distribution, while top-$k$ sampling \cite{fan2018hierarchical} restricts candidates to the $k$ most probable tokens, and nucleus (top-$p$) sampling \cite{holtzman2020curious} selects the smallest token set whose cumulative probability exceeds a threshold. These methods
trade output quality for diversity as their parameters increase, yet they share a fundamental limitation: all intervention occurs at the final vocabulary logit, leaving the model's internal computations, embeddings, attention patterns, and intermediate
representations entirely unchanged. Locally Typical Sampling \cite{meister2023locally} approaches diversity differently by selecting tokens whose information content stays close to the conditional entropy of the model, reducing degenerate repetition while preserving fluency. A complementary line of work intervenes on the output vocabulary itself: \citet{dineen2026vocabularydropoutcurriculumdiversity} randomly mask logits during self-play training to stop the model from collapsing onto a narrow set of token sequences. Their mask is applied during RL training, while ours is applied at inference, making the two approaches complementary. We use temperature, top-$p$, top-$k$ primary decoding baselines. Our internal noise methods are independent to all of them and can be combined with any output-level sampler.

\subsection{Internal Noise Injection in Transformers}

A smaller but closely related body of work explores injecting perturbations into the internal representations of transformer models rather than into the output distribution. \citet{kang2024lapael} is the most directly comparable prior work, they apply input-dependent Gaussian noise to the hidden states of early LLM layers during \textit{fine-tuning} to produce diverse paraphrases of training examples for
knowledge injection. \citet{kang2024lapael} demonstrate that latent-level perturbation yields more semantically diverse augmentations than data-level paraphrasing alone. This finding motivates our hypothesis that similar perturbations at \textit{inference time} can promote creative diversity in generation. Unlike \citet{kang2024lapael}'s method of applying noise during training to improve knowledge retention, our methods apply noise at inference time to diversify outputs without modifying model weights.

The residual stream framework of \citet{elhage2021mathematical} motivates our choice of injection sites. Embeddings, attention logits, and block outputs each represent distinct points at which information
is written to the model's shared communication channel, producing qualitatively different perturbation effects.

\subsection{Entropy-Aware Generation}

Several works use the model's own uncertainty as a signal for adaptive decoding. EDT \cite{zhang2024edt} raises temperature when the output vocabulary entropy is low, encouraging exploration precisely when the model is overconfident. Our attention entropy-based method shares this motivation but monitors entropy over attention distributions rather than output logits, intervening at the attention level, before the model's repetitive tendencies manifest in the output distribution.

\subsection{Constrained Generation and Arabic Educational LLMs}

Constrained decoding methods \cite{willard2023efficient} enforce formal output constraints by masking invalid tokens at each step, but do so at the cost of forcing the model toward low-probability and often repetitive continuations. This phenomena is seen clearly in Early-Grade Reading Assessment (EGRA) story generation \cite{rti2016egra, zualkernan2024EGRA}, where tight vocabulary and structural constraints narrow the valid token set and exacerbate the repetitive nature of outputs. Noise injection has been shown to systemically degrade a model's ability to follow constraints and guidelines \cite{shahani2025noise}, motivating careful calibration. Our noise decay strategy addresses this directly by reducing perturbation magnitude as generation progresses, allowing the model to explore creative directions early in the story while gradually restoring constraint following behaviour toward the conclusion. We evaluate story generation in this setting using a suite of small
Arabic-centric and multilingual instruct models \cite{sengupta2023jais,bari2024allam,huang2024acegpt,fanar2025}, measuring diversity via Self-BLEU \cite{zhu2018texygen} and semantic dispersion.

\section{Noise Steering methods}

\subsection{Embedding Noise}
Embedding noise injects Gaussian noise directly into the output of the token embedding lookup table during the decoding phase. This is the earliest possible injection site: the perturbed vector is the model's sole representation of the newly generated token before any transformer computation has occurred. All subsequent hidden-layer inputs are equivalent to the previous block's residual output and therefore have no independent embedding-level analogue.

Formally, let $\mathbf{e}_t \in \mathbb{R}^D$ denote the embedding of the token sampled at decode step $t$. We perturb it as:

\begin{equation}
    \tilde{\mathbf{e}}_t = \mathbf{e}_t + \varepsilon, \quad
    \varepsilon \sim \mathcal{N}\!\left(0,\, \sigma(t)^2 \mathbf{I}\right)
\end{equation}

where $\sigma(t) = \sigma_{\text{emb}} \cdot \delta(t)$ is the decayed standard deviation from Section~\ref{sec:decay}. Noise is applied exclusively to decode steps, the prefill pass over the prompt is left unperturbed, ensuring that the model's understanding of the instruction and constraints is not corrupted. Because the perturbation enters at the base of the residual stream, it is amplified and transformed by every subsequent attention and MLP layer, producing broad, high-level variation in the story's trajectory.

\subsection{Attention-logit Noise}

Attention-logit noise injects Gaussian noise into the output of the self-attention sub-layer at selected transformer blocks, specifically after the output projection $W_O$ but \emph{before} the subsequent MLP sub-layer and before the residual addition. Within each targeted block the computation is:

\begin{equation}
\mathbf{x} = \text{LayerNorm}(\mathbf{h}^{(l-1)})
\end{equation}

\begin{equation}
\mathbf{a} =
\operatorname{softmax}\!\left(
    \frac{QK^\top}{\sqrt{d}}
\right) V W_O + \varepsilon
\end{equation}

\begin{equation}
\mathbf{h}^{(l)} =
\mathbf{h}^{(l-1)} + \mathbf{a}
+ \operatorname{MLP}\!\left(
    \mathbf{a} + \mathbf{h}^{(l-1)}
\right)
\end{equation}

\begin{equation*}
\varepsilon \sim \mathcal{N}\!\left(
0, \sigma(t)^2 \mathbf{I}
\right)
\end{equation*}

where $\mathbf{h}^{(l)}$ is the residual stream at layer $l$ and $\sigma(t) = \sigma_{\text{attn}} \cdot \delta(t)$ follows the cosine decay schedule of Section~\ref{sec:decay}. The KV cache is left intact: the last token position of the attention output tensor (shape $(B, 1, D)$ during decoding) is perturbed only, leaving all cached key and value states unchanged.

By targeting the attention branch's contribution to the residual stream in isolation, before the MLP has processed the attended context, this method perturbs the information routing step of each selected block without corrupting the feed-forward transformation. This gives it a qualitatively different character from residual stream noise, which acts on the full block output after both sub-layers have run.

\subsection{Residual Stream Noise}

Residual stream noise injects Gaussian noise into the transformer's residual stream after each selected block has completed its full computation of both the self-attention and MLP sub-layers. The noise at layer $l$ and decode step $t$ is:
\[
    \tilde{h}^{(l)}_t = h^{(l)}_t + \varepsilon, \qquad \varepsilon \sim \mathcal{N}(0,\, \sigma(t)^2 \,)
\]
where $h^{(l)}_t$ is the block output at the last token position and $\sigma(t) = \sigma_{\mathrm{res}} \cdot \delta(t)$ is the decayed standard deviation from Section~\ref{sec:decay}. Noise is applied only during the decoding phase when new tokens are being generated, leaving the prompt representation unperturbed. Injecting at the full block output means the noise enters the residual
stream at the point where the block's complete contribution has been written, and propagates forward through all subsequent layers.

\subsection{Attention Entropy Noise Injection}

Attention Entropy Noise Injection (AENI) conditions noise magnitude on the peakedness of the model's own attention distribution at each decode step. The motivation, shared with EDT \cite{zhang2024edt}, is that a model is most likely to produce repetitive or formulaic output precisely when it is most confident: attention heads that concentrate their weight on a small set of tokens signal over-reliance on a narrow context window, which in constrained story generation often corresponds to copying earlier story elements. AENI responds by injecting more noise when attention is peaked and less when it is diffuse, intervening adaptively rather than applying a fixed perturbation schedule.

At each targeted layer $l$ and decode step $t$, we read the post-softmax
attention weight tensor $\mathbf{W}^{(l)} \in \mathbb{R}^{B \times H \times T_k}$, where $H$ is the number of heads and $T_k$ is the current key-sequence length. We extract the last query row (the current token's attention pattern)
$\mathbf{w} = \mathbf{W}^{(l)}_{:,:,-1,:}$ and compute a peakedness score as the mean per-head maximum attention weight:

\begin{equation}
    \phi(\mathbf{w}) = \operatorname{mean}_h \max_j\, w_{h,j}
\end{equation}

This score lies naturally in $[0,1]$, is independent of context length, and requires no normalisation: a peaked head drives its maximum toward 1, while a uniform distribution drives it toward $1/T_k$. Gaussian noise is then added to the self-attention output after $W_O$ but before the MLP and residual addition, with effective standard deviation:

\begin{equation}
    \sigma_{\text{eff}}(t) = \sigma_{\text{aeni}} \cdot \phi(\mathbf{w})
\end{equation}

where $\sigma_{\text{aeni}}$ is the per-model ceiling calibrated by RMS as in Section~\ref{sec:calibration}. As with attention-logit noise, only the last token position of the attention output (shape $(B, 1, D)$ during decoding) is perturbed and the KV cache is left intact.

\subsection{Noise Decay}
\label{sec:decay}
Applying constant noise throughout generation risks degrading constraint adherence, consistent with findings that activation-level noise
systemically harms instruction and constraint following in LLMs \cite{shahani2025noise}. We therefore apply a cosine decay schedule that reduces noise magnitude as generation progresses, allowing the model to explore creatively in the early tokens while recovering clean, constraint respecting decoding for the middle and conclusion of its story.

Let $t$ denote the current decode step and $T$ the decay horizon. The noise standard deviation at step $t$ is scaled by:
\[
    \delta(t) = \frac{1}{2}\left(1 + \cos\!\left(\frac{\pi \cdot \min(t,\, T)}{T}\right)\right)
\]
so that $\delta(0) = 1$ (full noise at the first generated token) and $\delta(T) = 0$
(no noise at and beyond the horizon), with the effective standard deviation given by
$\sigma(t) = \sigma_{\mathrm{(t-1)}} \cdot \delta(t)$.

\section{Experimental Setup}

\subsection{Noise Magnitude and Model Calibration}
\label{sec:calibration}
A key challenge when applying noise across multiple models is that the scale of internal representations varies substantially between architectures and models \cite{dettmers2022llmint8, sun2024massive}. A fixed noise standard deviation that is undetectable for one model may be catastrophically large for another. To ensure that noise is applied at a consistent \textit{relative} scale across all models, we calibrate the noise magnitude using the root mean square (RMS) of each model's residual stream activations.

For each model we register forward hooks on nearly all transformer blocks (excluding the first and last layers) and collect block output tensors during \textit{decode} steps only (i.e., when the sequence length is 1 and the KV cache is active), matching the phase at which noise is actually injected. The per-layer RMS is computed over the last generated token as:
\[
    \mathrm{RMS}(l) = \sqrt{\frac{1}{C} \sum_{C} x^2}
\]
where $x$ is the block output tensor at layer $l$ averaged over the batch and hidden dimensions, and $C$ is the number of decode steps collected. We then aggregate these per-layer values into a single scalar using the median across layers, which is robust to outlier layers at either extreme of the network.

The residual noise standard deviation for each model is then set as:
\[
    \sigma_{\mathrm{res}} = \alpha \cdot \mathrm{median}_l\bigl(\mathrm{RMS}(l)\bigr)
\]
where $\alpha$ is a fixed scaling coefficient set to $0.175$ across all experiments. This formulation ensures that the injected noise is always proportional to the model's own activation scale, making the effective perturbation strength comparable across models.

\subsection{Generation}
\label{sec:storygeneration}



We compare each noise-injection method to the following baselines: a \textit{noise-free} baseline ($T=1.0$), a \textit{High-temperature} ($T=1.8$) baseline with $topk=40$, and a \textit{High-temperature} ($T=1.8$) baseline with $topp=0.9$. The truncation values follow prior work: \citet{holtzman2020curious} report $k=40$ in their main comparison and note that nucleus values ``are usually in $[0.9, 1)$''. We pair these with an elevated $T=1.8$, within the high-temperature regime explored by \citet{nguyen2025turningheatminpsampling}. More aggressive settings (higher $T$, looser or no truncation) would likely produce still higher collapse rates, and absolute numbers would shift under per-model tuning. However, our principal claims concern the qualitative trade-off between output-level stochasticity and internal-representation perturbation, which we expect to be robust to such tuning.
 
Each condition generates 50 stories independently. Story $i$ under every condition is generated with the same seed, which makes generation reproducible and aligns the underlying sampler RNG state across conditions where the sampling distribution coincides. However, since different conditions modify the sampling distribution itself (e.g.\ via temperature, top-$k$, top-$p$, or internal noise), the seed alone does not eliminate sampling variability between conditions. We therefore report all aggregate metrics over the full 50-story sample per condition and use the statistical tests in Appendix~\ref{appendix:significance} to assess whether the observed effects are reliable rather than artefacts of sampling variance. All models are prompted with the same instruction to produce a story adhering to the EGRA constraints described in Section~\ref{sec:constraints}.

\subsection{Evaluation}

\begin{table*}[b]
\centering
\small
\setlength{\tabcolsep}{6pt}
\renewcommand{\arraystretch}{1.2}
\caption{%
  Modal collapse rate (\% of 50 generated stories) per model and decoding
  strategy.  \textbf{Italic}: method excluded from further evaluation due to (${\geq}\,40\%$ collapse rate);}
\label{tab:tmc}
\begin{tabular}{l *{7}{r}}
\toprule
\textbf{Model}
  & \textbf{Baseline}
  & \textbf{HiTemp-$k$}
  & \textbf{HiTemp-$p$}
  & \textbf{Embed}
  & \textbf{Attn}
  & \textbf{AENI}
  & \textbf{L-Res} \\
\midrule
ALLaM 7B
  & {0\%}
  & {0\%}
  & 4\%
  & {22}\%
  & {28}\%
  & {0\%}
  & {0\%} \\
 
AceGPT 8B
  & 2\%
  & \textit{84\%}
  & \textit{80\%}
  & {0\%}
  & {0\%}
  & {0\%}
  & 0\% \\
 
Fanar 9B
  & {0\%}
  & 2\%
  & {0\%}
  & {22}\%
  & {34}\%
  & 2\%
  & {0\%} \\
 
Jais 8B
  & {0\%}
  & {0\%}
  & {0\%}
  & 2\%
  & 4\%
  & {0\%}
  & {0\%} \\
 
Phi-4-mini
  & {0\%}
  & \textit{50\%}
  & \textit{66\%}
  & \textit{100\%}
  & \textit{44\%}
  & {0\%}
  & {0\%} \\
\bottomrule
\end{tabular}
\end{table*}

\subsubsection{Creativity Scores}
\label{sec:creativescore}
\paragraph{Vendi Score:} The \textit{Vendi Score} \cite{friedman2023vendiscorediversityevaluation} measures the diversity of a set of items by quantifying how evenly distributed they are across predefined categories. For our stories, embeddings were obtained using the \texttt{BAAI/bge-m3} model, with higher values indicating greater diversity.

\paragraph{Lexical Diversity:} We compute Self-BLEU \cite{zhu2018texygen}, treating each story in turn as a candidate and the remaining 49 as references. The lexical diversity score is $1 - \overline{\mathrm{Self\text{-}BLEU}}$, so that higher values indicate less overlap across the generated set.

\subsubsection{Readability and Linguistic Quality}
\label{sec:llm-judge}

We judge readability, grammatical correctness, linguistic quality, and appropriateness of reading level are evaluated using \textsc{GPT5.3 Chat} \cite{openai2026gpt53instant} as an LLM judge. Each story is scored independently on these dimensions, providing a complementary
quality signal to the diversity metrics above.

\subsubsection{Constraint Following}
\label{sec:constraints}

EGRA constraint adherence is evaluated through a combination of rule-based checks and LLM judgement. Constraints that are able to be verified exactly are checked programmatically. These constraints include: Story length not exceeding 60 words, Use of exactly one proper noun, Use of present tense

Constraints requiring semantic and structural understanding are evaluated using \textsc{GPT5.3 CHAT} \cite{openai2026gpt53instant} as a judge. These constraints include:\\
- Narrative structure includes intro, middle dilemma, and ending with resolution.\\
- Gender-balanced: includes both a boy and a girl.\\
- Avoids gender/religion/other stereotypes.\\
- Vocabulary suitable for children and local context.

\subsection{Models and Metrics}
\label{sec:models-metrics}

\textbf{Models: }We evaluate five small ($\leq$9B parameter) Arabic-centric instruct models. \textbf{Jais 8B} \cite{sengupta2023jais} is pretrained from scratch on a large Arabic and English corpus. \textbf{ALLaM 7B} \cite{bari2024allam} adapts LLaMA via Arabic tokeniser expansion and continued pretraining. \textbf{Fanar 9B} \cite{fanar2025} emphasises morphology-aware Arabic tokenisation and dialectal coverage. \textbf{AceGPT} \cite{huang2024acegpt} is a bilingual (Arabic--English) model built on the LLaMA architecture with continued pretraining on large-scale Arabic data and instruction tuning for dialogue and generation tasks. \textbf{Phi-4-mini} is included as a compact multilingual baseline. Together these models span the main architectural and training strategies present in the current Arabic SLM landscape.

\paragraph{Quality (Qual.\,$\uparrow$).} The quality score is an average of four LLM-judge dimensions assessed per story: overall readability, logical soundness, grammatical correctness, linguistic quality. Each dimension is scored independently out of 10 and the four scores are averaged into a single value per story, which is then averaged across the 50 stories in a condition.

\paragraph{Constraint Violations (Viol.\,$\downarrow$).}
We report the average number of constraints broken per story and the total constraints broken, where a lower score is better and zero indicates full compliance. This combines the programmatic and LLM-judge checks from Section~\ref{sec:constraints}: each of the seven constraints (three programmatic, four LLM-judged) contributes one violation point if failed, giving a maximum possible score of 7 per story.

\section{Results}

\subsection{Total Modal Collapse}
\label{sec:TMC}

 \begin{figure*}[b]
    \centering
    \includegraphics[width=1\linewidth]{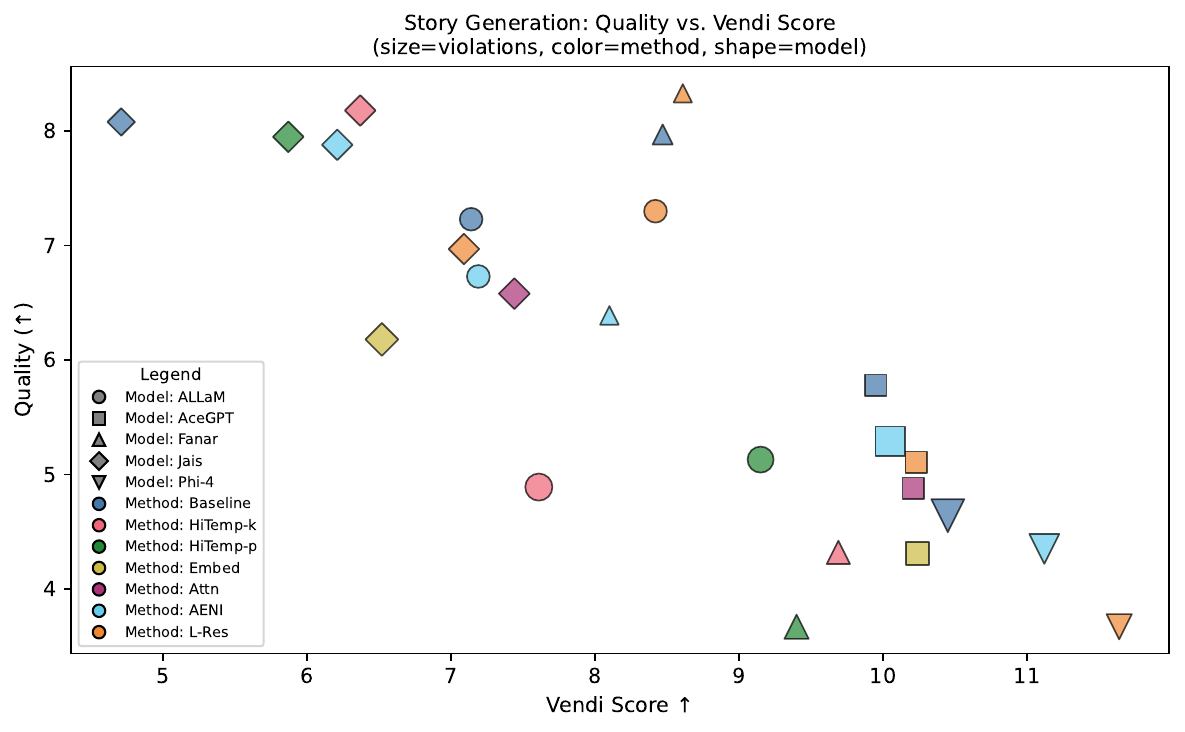}
    \caption{Story quality versus output diversity (Vendi Score) for all evaluated conditions with $\leq$ 10 total modal collapses. Each point represents one model–method combination: marker \textbf{shape} encodes the model and marker \textbf{colour} encodes the decoding method; marker \textbf{area} is proportional to the total number of EGRA constraint violations across the 50 generated stories (larger = more violations). Methods that cluster toward the lower-left or produce large markers trade quality or diversity for higher violation rates. Conditions with > 10 modal collapses are excluded to avoid diversity scores inflated by repetitive collapse outputs}
    \label{fig:scatter}
\end{figure*}

Before comparing diversity and quality metrics, we first examine instances of Total modal collapse (TMC): cases where generation degenerates into repetitive or incoherent token sequences, producing stories that are effectively unusable. Table~\ref{tab:tmc} reports the collapse rate for each model and decoding condition. Any condition with a collapse rate at or above 40\% is excluded from further evaluation, as the surviving outputs would constitute a non-representative sample. Conditions with collapse rates greater than 20\% (and less than 40\%) are excluded from Figure \ref{fig:scatter} but are discussed further in Sec \ref{sec:mainResults}.

\paragraph{High-temperature baselines.}The high temperature conditions (HiTemp-k and HiTemp-p) are the most fragile overall. AceGPT and Phi-4-mini both exceed the 40\% threshold under both high-temperature settings, with collapse rates as high as 84\% and 66\% respectively. This confirms the well-known instability of aggressive output-level sampling: flattening the output distribution beyond a certain point causes the model to assign non-negligible probability to incoherent continuations, from which recovery is rare once the context has been corrupted \cite{nguyen2025turningheatminpsampling}.

\paragraph{Residual stream noise.} residual stream noise (\textsc{L-Res}), records a 0\% collapse rate across every model tested. Because this method injects noise after each transformer block has already completed its full computation, the perturbation enters the network at a point where the model's internal representations are well-formed, and the signal propagates forward in a controlled manner. This makes residual noise a reliable location to inject noise without worrying about generation failures.

\paragraph{Attention noise and AENI.} Raw attention-logit noise (\textsc{Attn}) shows high collapse rates, reaching 28\% and 34\% on ALLaM and Fanar respectively. However, our Attention Entropy Noise Injection method (\textsc{AENI}) eliminates collapse entirely across all five models. Rather than applying a fixed perturbation at every step, AENI scales the noise magnitude according to how peaked the model's own attention distribution is at that moment. When the model attends broadly and diffusely, little or no noise is injected; perturbation is reserved for steps where attention has collapsed onto a narrow set of tokens, which is precisely when formulaic or repetitive output is most likely to follow. This adaptive gating prevents the runaway instability that fixed attention noise produces.

\paragraph{Embedding noise.} Embedding noise (\textsc{Embed}) causes complete collapse on Phi-4-mini (100\%), and elevated rates on ALLaM (22\%) and Fanar (22\%). The Phi-4-mini model in particular is very sensitive to embedding noise, even at noise levels $\alpha < 0.175$ virtually every generation collapses. As a compact multilingual model, Phi-4-mini has had comparatively less Arabic training data, meaning its token embedding space for Arabic is more sparsely and less robustly organized. Because embedding noise acts at the very first step of the residual stream, before any transformer computation has occurred, any perturbation there propagates through every subsequent layer. 

For a model whose Arabic representations are already fragile, this earliest injection site offers no opportunity for the model to compensate.

\subsection{Diversity, Quality, and Constraint Adherence}
\label{sec:mainResults}

Figure~\ref{fig:scatter} plots each model-method condition in the space of Vendi Score (diversity, x-axis) against LLM-judged quality (y-axis), with marker size encoding the average number of constraint violations per story. Beyond these core metrics, we also track estimated reading grade level per condition, as diversity methods that push stories toward richer vocabulary or more complex sentence structures may improve narrative variety and quality scores while silently violating the early-grade readability that EGRA passages require. Full numerical results are in Appendix Tables~\ref{table:quality} and~\ref{tab:osman}.

\paragraph{Residual Stream Noise.}
\textsc{L-Res} is the most consistently beneficial intervention. It achieves a 0\% collapse rate for every model and improves Vendi Score in all five cases: +1.28 for ALLaM (7.14→8.42), +0.28 for AceGPT (9.95→10.23), +0.14 for Fanar (8.47→8.61), +2.38 for Jais (4.71→7.09), and +1.19 for Phi-4 (10.45→11.64). Constraint adherence is largely undisturbed, with violations changing by at most 0.78 per story across all models. Reading grade level is preserved throughout: ALLaM and Fanar remain at Grade 3, Jais at Grade 2, and AceGPT at Grade 4, all consistent with their respective baselines. These results indicate that perturbing the model's internal representations allows it to diversify narrative choices, such as character, setting, and plot structure, without reaching for more complex language to do so. The quality impact depends on the baseline. For Fanar and ALLaM, which already produce strong stories (quality scores of 7.97 and 7.23 respectively), \textsc{L-Res} maintains or improves quality outright: Fanar reaches 8.33 while gaining diversity, placing it in a genuinely better region of the quality-diversity space. For weaker models the diversity gains come at a greater quality cost: AceGPT drops from 5.78 to 5.11, Jais from 8.08 to 6.97, and Phi-4 from 4.64 to 3.67. In Jais's case the large Vendi Score gain suggests the model was simply not exploring the narrative space at baseline. In Phi-4's case, even gentle perturbation strains its more limited Arabic generation capability, and grade level drifts to Grade 5, the one exception to \textsc{L-Res}'s otherwise consistent grade-level preservation.

\paragraph{Attention Entropy Noise Injection vs. Raw Attention Noise.}
Raw attention-logit noise (\textsc{Attn}) can achieve high diversity, with Vendi Scores of 9.51 for ALLaM and 7.44 for Jais, but at severe cost: collapse rates of 28\% and 34\% on ALLaM and Fanar respectively, constraint violations spiking to 5.36, and ALLaM quality collapsing to 2.55, the lowest value in the entire evaluation result in the surviving scores being a biased subset of an otherwise unreliable condition. \textsc{AENI} resolves this by conditioning noise magnitude on the model's own attention entropy, injecting more perturbation when attention is sharply peaked and backing off when it is already diffuse. Collapse rates fall to 0\% on ALLaM, AceGPT, and Jais, and just 2\% on Fanar, with quality recovering substantially: ALLaM to 6.73, Fanar to 6.39, and Jais to 7.88. Crucially, \textsc{AENI} also preserves reading grade level across all Arabic-centric models and for Phi-4 actually improves it to Grade 3 from the baseline Grade 4, a stronger result than \textsc{L-Res} on that model. The cost of this stability is a moderate Vendi Score reduction relative to raw \textsc{Attn}, but this trade-off is clearly worthwhile for an educational generation system. On AceGPT, already a high-diversity model, the two methods produce near-identical Vendi Scores (10.21 vs. 10.05), confirming that \textsc{AENI}'s adaptive mechanism does not sacrifice diversity unnecessarily when it is not needed.

\paragraph{High-temperature Baselines} High-temperature sampling is effective only for Jais, whose baseline Vendi Score of 4.71 is the lowest in the evaluation. Both \textsc{HiTemp-k} and \textsc{HiTemp-p} raise diversity meaningfully (to 6.37 and 5.87) while quality stays high at 8.18 and 7.95, suggesting Jais's cautious baseline was due to overfitting to the EGRA task format, producing structurally valid but narratively repetitive stories, and that output-level stochasticity was sufficient to break this pattern without destabilising the underlying generation quality. For every other model the picture is far worse: ALLaM's quality falls to 4.89 and 5.13, Fanar's to 4.32 and 3.67, and both AceGPT and Phi-4 exceed the collapse threshold entirely. The grade-level results are equally damaging: ALLaM's Grade 3 baseline rises to Grade 6 and Grade 5 under the two conditions, and Fanar's Grade 3 baseline drifts to Grade 5 and Grade 6 respectively. By flattening the output probability distribution, high-temperature sampling makes models more likely to reach for less common, more sophisticated vocabulary, precisely the kind of variation that is undesirable in an early-grade assessment context. It treats all forms of output variation as equivalent and has no mechanism for distinguishing a more creative plot from a more adult vocabulary, a distinction that is central to the EGRA task.

\paragraph{}
Appendix~\ref{appendix:significance} contains significance tests that re-affirm these findings: L-RES and AENI are the only methods that consistently preserve quality and constraint adherence at a level statistically indistinguishable from the Baseline, while high-temperature sampling, embedding noise and raw attention noise injections are significantly worse on both metrics across the majority of models.

\section{Conclusion}

We investigated noise steering as a lightweight, fine-tuning-free approach to improving narrative diversity in Arabic EGRA-style story generation across five small language models. \textsc{L-RES} and \textsc{AENI} are the only methods that consistently improve diversity without the substantial degradation in quality, constraint adherence, or early-grade readability that the alternatives produce, the last of which proves a critical differentiator from high-temperature sampling, which inflates reading level and causes catastrophic collapse on several models. Residual stream noise is the most reliable all-around method and yields the greatest gains on models that already generate high-quality stories at baseline. \textsc{AENI} trades a small diversity margin for substantially better stability, making it the stronger choice when attention-level intervention is preferred. A consistent finding across all methods is that a model's baseline generation quality is a strong predictor of how gracefully it absorbs noise: models with weaker Arabic representation degrade faster regardless of the intervention applied, suggesting that noise steering is best understood as a tool for unlocking latent diversity rather than compensating for limited language capability. Future work should explore layer-specific targeting strategies and extension to other low-resource languages and assessment frameworks.

\section*{Limitations}

\paragraph{LLM-as-a-Judge Evaluation.}
Our quality and parts of our constraint adherence scores rely on GPT as an LLM judge, which introduces well-documented limitations. LLM judges are known to exhibit systematic biases including verbosity bias, where longer outputs receive inflated scores, and self-enhancement bias, where a model tends to favor outputs stylistically similar to its own generations \cite{chen-etal-2024-humans}. A particular concern in comparative evaluations is relative scoring bias, where a story may receive an inflated score simply by virtue of being better than the others it is presented alongside rather than on absolute merit. We took two steps to mitigate this: stories were submitted to the judge in fixed-size batches rather than all at once, and stories from different experimental conditions were shuffled together before evaluation, preventing the judge from implicitly comparing outputs within a single run. Quality dimensions such as narrative clarity and vocabulary suitability ideally require human annotators with relevant expertise. The programmatic constraint checks mitigate this concern for verifiable constraints such as word count and tense, but the LLM-judged dimensions: narrative structure, cultural neutrality, and reading appropriateness, should be interpreted with caution. Future work should incorporate human evaluation by Arabic literacy educators to validate the automated scores reported here.

\paragraph{Decoding-Hyperparameter Sensitivity.}
Our high-temperature baselines use single fixed values of $T = 1.8$, $k = 40$, and $p = 0.9$, chosen as widely used high-diversity defaults in prior work but not separately tuned per model. Since different models exhibit different baseline confidence profiles, the absolute diversity, quality, and collapse numbers for these baselines could shift under per-model hyperparameter sweeps. A similar caveat applies to the single value of $\alpha = 0.175$ used to calibrate noise magnitude across all noise-injection methods. We expect the qualitative ordering between output-level stochastic decoding and internal-representation perturbation to be robust, but a more comprehensive sensitivity study across decoding hyperparameters and per-model $\alpha$ values is left to future work.

\section*{Acknowledgments}

This work was funded by the American University of Sharjah through Faculty Research Grant FRG24-E-E87.

\bibliography{custom}

\newpage
\appendix
\section{Statistical Significance: L-Res and AENI vs.\ Alternatives}
\label{appendix:significance}

\paragraph{Test choice.}
We compare each method's per-story quality and constraint-violation distributions to the Baseline using the Kruskal-Wallis omnibus test followed by Dunn's pairwise post-hoc with Holm correction. This single non-parametric procedure is the appropriate choice for our data: Levene's test rejects equality of variances ($p < 0.05$) in 7 of 10 (model, metric) groups, ruling out ANOVA + Tukey HSD; the metrics are also bounded and non-normal (quality $\in [0,10]$ with TMC zero-mass; violations integer counts $\in [0,7]$). Effect size is the matching non-parametric statistic, the rank-biserial correlation $r \in [-1, +1]$, with the convention that $r > 0$ indicates the method is \emph{better than the Baseline} on that metric (quality higher, violations lower); $|r| < 0.1$ is negligible, $\geq 0.3$ medium, $\geq 0.5$ large, $|r| = 1$ is the maximum the metric supports.

\paragraph{Every alternative method significantly degrades performance whenever it can be tested.} Tables~\ref{tab:dunn-q} and~\ref{tab:dunn-v} report rank-biserial $r$ between each method and the Baseline for quality and violations respectively. Counting cells where the omnibus K-W is significant and the method survived TMC exclusion (the cells where Dunn's post-hoc is licensed to run), the asymmetry between our proposed methods and the alternatives is severe: \textsc{Embed} produces a significant adverse deviation in \textbf{every single testable cell (6 of 6)};
\textsc{Attn} in 5 of 6; \textsc{HiTemp-$k$} and \textsc{HiTemp-$p$} in 3 of 5 each. By contrast, \textsc{L-Res} and \textsc{AENI} each produce a
significant adverse deviation in only 2 of 7 testable cells. Where the alternatives register a significant effect, the median $|r|$ is $0.65$ and \emph{three separate cells reach $|r| = 0.99$, the maximum the metric supports}; where L-Res and AENI register one, $|r|$ never exceeds $0.48$, and the median is $0.40$. There is no model and no metric on which any alternative method is significantly closer to the Baseline than L-Res or AENI.

\begin{table}[h]
\centering
\small
\setlength{\tabcolsep}{4pt}
\renewcommand{\arraystretch}{1.15}
\caption{%
\textbf{Quality.} Rank-biserial $r$ between each method and the
Baseline. Negative $r$ indicates worse quality than Baseline.
\textbf{Bold}: Dunn's $p\!<\!.05$ (Holm-adjusted), with stars
$^{*}p\!<\!.05$, $^{**}p\!<\!.01$, $^{***}p\!<\!.001$.
``---'': condition excluded by 40\% TMC threshold.
The Phi-4-mini omnibus K-W is non-significant ($H\!=\!4.2$,
$p\!=\!.12$); the column is shown for completeness but no Dunn's
test is licensed.
Other omnibus K-W tests: ALLaM $H\!=\!159.8$,
AceGPT $H\!=\!32.8$, Fanar $H\!=\!158.9$, Jais $H\!=\!62.8$
(all $p\!<\!.001$).}
\label{tab:dunn-q}
\begin{tabular}{l c c c c c}
\toprule
\textbf{Method} & ALLaM & AceGPT & Fanar & Jais & Phi-4 \\
\midrule
\textsc{L-Res} & $-.02$ & $-.30$ & $+.26$ & $\mathbf{-.43^{**}}$  & $-.22$ \\
\textsc{AENI}  & $-.39$ & $-.28$ & $\mathbf{-.48^{*}}$  & $-.06$ & $-.18$ \\
\midrule
\textsc{Attn}  & $\mathbf{-.99^{***}}$ & $\mathbf{-.46^{**}}$  & $\mathbf{-.67^{***}}$ & $\mathbf{-.48^{***}}$ & --- \\
\textsc{Embed} & $\mathbf{-.62^{***}}$ & $\mathbf{-.61^{***}}$ & $\mathbf{-.75^{***}}$ & $\mathbf{-.57^{***}}$ & --- \\
HiTemp-$k$     & $\mathbf{-.64^{***}}$ & ---                   & $\mathbf{-.81^{***}}$ & $+.01$                & --- \\
HiTemp-$p$     & $\mathbf{-.65^{***}}$ & ---                   & $\mathbf{-.99^{***}}$ & $-.06$                & --- \\
\bottomrule
\end{tabular}
\end{table}

\begin{table}[h]
\centering
\small
\setlength{\tabcolsep}{4pt}
\renewcommand{\arraystretch}{1.15}
\caption{%
\textbf{Constraint Violations.} Rank-biserial $r$ between each method
and the Baseline. Negative $r$ indicates more violations than
Baseline. \textbf{Bold}: Dunn's $p\!<\!.05$ (Holm-adjusted).
``---'': condition excluded by 40\% TMC threshold. AceGPT and Jais
omnibus K-W are non-significant ($p\!=\!.19$ and $p\!=\!.08$); those
columns show $r$ values for completeness but no Dunn's test is
licensed. Other omnibus K-W tests: ALLaM $H\!=\!80.9$, Fanar
$H\!=\!31.3$, Phi-4-mini $H\!=\!13.3$ (all $p\!<\!.01$).}
\label{tab:dunn-v}
\begin{tabular}{l c c c c c}
\toprule
\textbf{Method} & ALLaM & AceGPT & Fanar & Jais & Phi-4 \\
\midrule
\textsc{L-Res} & $+.04$ & $-.07$ & $-.06$ & $-.23$ & $\mathbf{-.32^{**}}$ \\
\textsc{AENI}  & $+.12$ & $-.08$ & $-.22$ & $-.32$ & $\mathbf{-.38^{**}}$ \\
\midrule
\textsc{Attn}  & $\mathbf{-.69^{***}}$ & $-.08$ & $-.08$               & $-.20$ & --- \\
\textsc{Embed} & $\mathbf{-.50^{***}}$ & $-.25$ & $\mathbf{-.33^{**}}$ & $-.33$ & --- \\
HiTemp-$k$     & $\mathbf{-.34^{*}}$   & ---    & $-.30$               & $-.21$ & --- \\
HiTemp-$p$     & $-.25$                & ---    & $\mathbf{-.51^{***}}$& $-.22$ & --- \\
\bottomrule
\end{tabular}
\end{table}

\paragraph{When alternatives degrade quality, they degrade it catastrophically.} The visual pattern in Table~\ref{tab:dunn-q} is unambiguous: every significant deviation produced by \textsc{Attn}, \textsc{Embed}, \textsc{HiTemp-$k$}, or \textsc{HiTemp-$p$} on quality is large or catastrophic ($|r| \geq 0.46$, with seven of fourteen at $|r| \geq 0.65$ and three at $|r| = 0.99$, where the metric saturates). The corresponding L-Res and AENI cells average $|r| = 0.20$, well below the medium-effect threshold, and on several models L-Res sits within $|r| < 0.05$ of the Baseline distribution, indistinguishable from no intervention at all. The same pattern holds for violations (Table~\ref{tab:dunn-v}): the only large-magnitude deviation ($\mathbf{-.69^{***}}$ for \textsc{Attn} on ALLaM) belongs to an alternative method, and the only $|r| \geq 0.5$ result outside L-Res and AENI's row is HiTemp-$p$ on Fanar.

\paragraph{The L-Res and AENI exceptions.} The two cells per method on which L-Res or AENI register a significant deviation each correspond to a known and bounded trade-off. \textsc{L-Res} on Jais quality ($r\!=\!-.43$, $p\!=\!.002$) is the model with the largest single diversity gain in the entire study (Vendi $4.71 \to 7.09$); a medium quality effect there is the cost of the largest narrative-diversity improvement we report. \textsc{AENI} on Fanar quality ($r\!=\!-.48$, $p\!=\!.029$) sits well below the catastrophic $r\!=\!-.67$ that raw \textsc{Attn} produces on the same model, showing that AENI's adaptive gating recovers most of the quality that fixed attention noise destroys. Both methods also register a medium adverse effect on Phi-4-mini violations ($r\!=\!-.32$ and $-.38$); however, on this model \textsc{L-Res} and \textsc{AENI} are the \emph{only two methods that survive TMC exclusion at all}. \textsc{Embed}, \textsc{Attn}, \textsc{HiTemp-$k$}, and \textsc{HiTemp-$p$} all collapse on $\geq 40\%$ of generations and are excluded entirely from this analysis. The choice on Phi-4-mini is therefore between a medium increase in violations under L-Res or AENI, and complete generation collapse under every alternative tested.

\newpage

\section{Story Generation Prompt}

System Prompt:
\begin{quote}
\begin{Arabic}
أنتَ مساعدٌ مُفيدٌ تُعدّ نصوص قراءةٍ مُخصصةٍ للأطفال الصغار لتنمية مهاراتهم في فهم المقروء.
\end{Arabic}
\end{quote}

\textcolor{white}{|}\\

English Translation of System Prompt:
\begin{quote}
You are a helpful assistant who prepares reading texts specifically for young children to develop their reading comprehension skills.
\end{quote}

\textcolor{white}{|}\\

User Prompt:
\begin{quote}
\begin{Arabic}
اكتب قصه.

* يجب أن تكون القصة سردية مستوحاة من مواد قراءة الأطفال، وتتضمن:

* مقدمة تُعرّف بالشخصيات

* جزءًا وسطيًا يتضمن معضلة ما

* جزءًا ختاميًا يتضمن حدثًا لحل المعضلة

* يجب ألا تتجاوز القصة 60 كلمة.

* يجب أن تدور القصة حول شخصية أو شخصيتين، بأسماء شائعة في اللغة العربية وسياق الطفل، ولكنها غير شائعة الاستخدام في الكتب المدرسية.

* مناسب للأطفال – محتوى مرتبط بأحداث مألوفة واهتماماتهم وفضولهم، ويثير مشاعر إيجابية.

* يحتوي على عناصر القصة القصيرة: شخصية، سياق، بداية، عقبة أو مشكلة، وحل.

* متوازن بين الجنسين – يضم كلاً من الأولاد والبنات.

* يتجنب الصور النمطية المتعلقة بالجنس أو الدين أو غيرها.

* لا يوجد نص موجود مسبقًا ولا يذكر الأطفال بقصص أو أساطير يعرفونها.

* يستخدم زمن المضارع.

* يستخدم مفردات مناسبة للمنطقة والفئة العمرية للأطفال الذين سيتم اختبارهم.

* يجب أن تكون الجملة الأولى سهلة للغاية.
* يستخدم بنية متنوعة ولكنها ليست أدبية أو معقدة.

* يستخدم اسمًا علمًا واحدًا (شائعًا) فقط.

* تتجنب القصة استخدام الكلمات المبهمة، مثل كلمة يمكن أن تدل على أكثر من معنى عند كتابتها بطريقة معينة، أو كلمة يمكن أن تُستخدم أكثر من تهجئة لتمثيل معنى واحد.

* ليس قائمة من الجمل المترابطة بشكل ضعيف.

* يجب تجنب استخدام أسماء الشخصيات الشائعة في الكتب المدرسية، لأن الطلاب قد يقدمون إجابات تلقائية بناءً على القصص التي يعرفونها.

* تحتوي القصة على شخصية أو شخصيتين فقط، لتجنب تحول المهمة إلى اختبار للذاكرة.

* يحتوي نص القصة على بعض المفردات المعقدة وتراكيب الجمل.

اكتب القصة كنص سردي متصل دون عناوين أو تسميات أو تقسيمات أو أي مؤشرات هيكلية.
\end{Arabic}
\end{quote}

\textcolor{white}{|}\\

English Translation of Constraints:
\begin{quote}
The story must be a narrative story generated from children’s reading material which has a beginning section where the characters are introduced, a middle section containing some dilemma, and an ending section with an action resolving the dilemma.

The story should be 60 words long.

The story should revolve around one to two characters with names that are common to the Arabic language and context of the child but not commonly used in school textbooks.

Appropriate for children - content related to familiar events, their interests, and their curiosity and evokes positive emotions

Has the elements of a short story: a character, context, beginning, obstacle or problem, and a resolution

Gender balanced – feature both boys and girls

Avoids gender, religious or other stereotypes

Does not already exist or remind children of stories or legends they already know

Uses the present tense

Uses vocabulary that is appropriate to the region and age of the children to be tested

The first sentence should be very easy

Uses varied structure but is not too literary/complicated
Allows for a variety of comprehension questions (literal and inferential)

Only uses one (common) proper name

The story avoids the use of ambiguous words, such as a word that, spelled in one way, can represent more than one meaning or a word that can use more than one spelling to represent one meaning

It is not a list of loosely connected sentences.

Character names frequently used in the school textbook are to be avoided, as students may give automated responses based on the stories with which they are familiar

The story has only one to two characters, to avoid the task becoming about memory recall

The story text contains some complex vocabulary and sentence structures.

\end{quote}

\newpage
\section{LLM-as-a-judge prompt}

The following prompt was given to \textsc{GPT5.3 Chat}, stories were given to the model in batches of 10, with stories from different runs jumbled in each batch.

\begin{lstlisting}[caption={Judging prompt fed to \textsc{GPT5.3 Chat}}]
The following are a set of Arabic stories. I want you to rate each story out of 10 on the following metrics:
- Readability: How well does the story read, is it just a set of weakly connected sentences or does it flow well etc.
- Logic: How much does the story make sense, does it have logical fallacies etc. - Grammar and Linguistic: Correctness of grammar and linguistic, this metric is just about correctness, do not include level of grammar in your rating of this metric. Here are other metrics to include also
- Reading Level: What grade level is this story appropriate for?
- Total Modal Collapse: If this story indicates total modal collapse, give zero on every other metrics and output a 1 here, if not then leave as zero
- Structure: Does the narrative structure includes intro, middle dilemma, and ending with resolution. 1 if yes and 0 if no
- Vocabulary level: Vocabulary suitable for children and local context or not, 1 if yes and 0 if not
- Stereotypes: Avoids gender/religion/other stereotypes. 1 if yes and 0 if not
- Gender-balanced: includes both a boy and a girl. 1 if yes and 0 if not


Your response should be as a table format csv with the following column names:
Story number, Readability, Logic, GrammarandLinguistics, ReadingLevel, TotalModalCollapse, Structure, VocabularyLevel, Stereotypes, Gender-balanced"
\end{lstlisting}

\newpage

\section{Readability and Grade Level Scores}
\label{sec:osman}

\begin{table*}[t]
\centering
\small
\setlength{\tabcolsep}{5pt}
\renewcommand{\arraystretch}{1.15}
\caption{%
  Readability and constraint-violation metrics for all conditions with
  ${<}\,40\%$ modal collapse.
  \textbf{Read.}~$\uparrow$: mean OsmanReadability score \cite{el-haj-rayson-2016-osman}~$\pm$\,std
  (higher = easier to read).
  \textbf{Grade}: modal estimated grade level across generated stories
  (G1--G3 = EGRA target range; higher grades indicate above-target
  reading difficulty; \textit{0} = output mostly unreadable).
  \textbf{Viol.}~$\downarrow$: mean EGRA constraint violations per
  story~$\pm$\,std.
  \textbf{Bold}: best readability and fewest violations among methods
  with ${\leq}\,1$ collapse per model group.
  \textsuperscript{$\dagger$}~method has ${\geq}\,1$ collapse
  (count in Table~\ref{tab:tmc}).}
\label{tab:osman}
\begin{tabular}{l l c c c}
\toprule
\textbf{Model} & \textbf{Method}
  & \textbf{Read.}~$\uparrow$
  & \textbf{Grade}
  & \textbf{Viol.}~$\downarrow$ \\
\midrule
\multirow{7}{*}{ALLaM 7B}
  & Baseline
    & \textbf{70.42} $\pm$ 9.40  & G3 & 3.38 $\pm$ 0.67 \\
  & HiTemp-$k$
    & 62.90 $\pm$ 20.30 & G6 & 4.40 $\pm$ 1.55 \\
  & HiTemp-$p$\textsuperscript{$\dagger$}
    & 76.57 $\pm$ 21.89 & G5 & 4.10 $\pm$ 1.42 \\
  \cmidrule(l){2-5}
  & \textsc{Embed}\textsuperscript{$\dagger$}
    & 59.76 $\pm$ 117.64 & \textit{0} & 4.92 $\pm$ 1.44 \\
  & \textsc{Attn}\textsuperscript{$\dagger$}
    & 74.07 $\pm$ 32.86  & \textit{0} & 5.36 $\pm$ 1.57 \\
  & \textsc{AENI}
    & 39.78 $\pm$ 211.51 & G3 & \textbf{3.22} $\pm$ 0.42 \\
  & \textsc{L-Res}
    & 76.81 $\pm$ 10.86  & G3 & 3.40 $\pm$ 0.86 \\
\midrule
\multirow{5}{*}{AceGPT 8B}
  & Baseline\textsuperscript{$\dagger$}
    & 82.12 $\pm$ 6.59  & G4 & 3.08 $\pm$ 1.05 \\
  \cmidrule(l){2-5}
  & \textsc{Embed}
    & \textbf{85.84} $\pm$ 7.29  & G3 & 3.54 $\pm$ 1.13 \\
  & \textsc{Attn}
    & 85.02 $\pm$ 6.39  & G3 & {3.12} $\pm$ 0.87 \\
  & \textsc{AENI}
    & 81.84 $\pm$ 9.43  & G4 & 3.14 $\pm$ 0.88 \\
  & \textsc{L-Res}
    & 85.48 $\pm$ 11.85 & \textit{G4} & \textbf{3.10} $\pm$ 0.84 \\
\midrule
\multirow{7}{*}{Fanar 9B}
  & Baseline
    & \textbf{79.25} $\pm$ 6.33  & G3 & \textbf{2.86} $\pm$ 0.57 \\
  & HiTemp-$k$\textsuperscript{$\dagger$}
    & 52.65 $\pm$ 16.99 & G5 & 3.48 $\pm$ 1.11 \\
  & HiTemp-$p$
    & 53.55 $\pm$ 17.34 & G6 & 3.83 $\pm$ 1.14 \\
  \cmidrule(l){2-5}
  & \textsc{Embed}\textsuperscript{$\dagger$}
    & 42.94 $\pm$ 66.52  & \textit{0} & 4.42 $\pm$ 1.94 \\
  & \textsc{Attn}\textsuperscript{$\dagger$}
    & 82.95 $\pm$ 5.91   & \textit{0} & 3.54 $\pm$ 1.49 \\
  & \textsc{AENI}\textsuperscript{$\dagger$}
    & 80.56 $\pm$ 5.81   & G3 & 3.32 $\pm$ 1.08 \\
  & \textsc{L-Res}
    & 77.97 $\pm$ 5.26   & G3 & 2.90 $\pm$ 0.46 \\
\midrule
\multirow{7}{*}{Jais 8B}
  & Baseline
    & \textbf{85.63} $\pm$ 7.62 & G2 & \textbf{2.74} $\pm$ 0.94 \\
  & HiTemp-$k$
    & 84.21 $\pm$ 7.48  & G2 & 3.18 $\pm$ 1.10 \\
  & HiTemp-$p$
    & 84.85 $\pm$ 7.55  & G2 & 3.18 $\pm$ 1.17 \\
  \cmidrule(l){2-5}
  & \textsc{Embed}\textsuperscript{$\dagger$}
    & 34.67 $\pm$ 177.19 & G2 & 3.46 $\pm$ 1.07 \\
  & \textsc{Attn}\textsuperscript{$\dagger$}
    & 78.63 $\pm$ 41.50  & G2 & 3.18 $\pm$ 1.02 \\
  & \textsc{AENI}
    & 85.45 $\pm$ 6.12   & G2 & 3.32 $\pm$ 0.89 \\
  & \textsc{L-Res}
    & 82.93 $\pm$ 10.13  & G2 & 3.20 $\pm$ 1.04 \\
\midrule
\multirow{3}{*}{Phi-4-mini}
  & Baseline
    & \textbf{82.02} $\pm$ 6.30 & G4 & \textbf{3.16} $\pm$ 1.02 \\
  \cmidrule(l){2-5}
  & \textsc{AENI}
    & 80.59 $\pm$ 15.23 & G3 & 3.96 $\pm$ 1.16 \\
  & \textsc{L-Res}
    & 82.37 $\pm$ 7.50  & G5 & 3.94 $\pm$ 1.42 \\
\bottomrule
\end{tabular}
\end{table*}

Table \ref{tab:osman} contains OsmanReadability Scores \cite{el-haj-rayson-2016-osman} that were calculated to go along with the Grade level assessments made from our LLM-as-a-judge model.

\begin{figure*}[t]
    \centering
    \includegraphics[width=\linewidth]{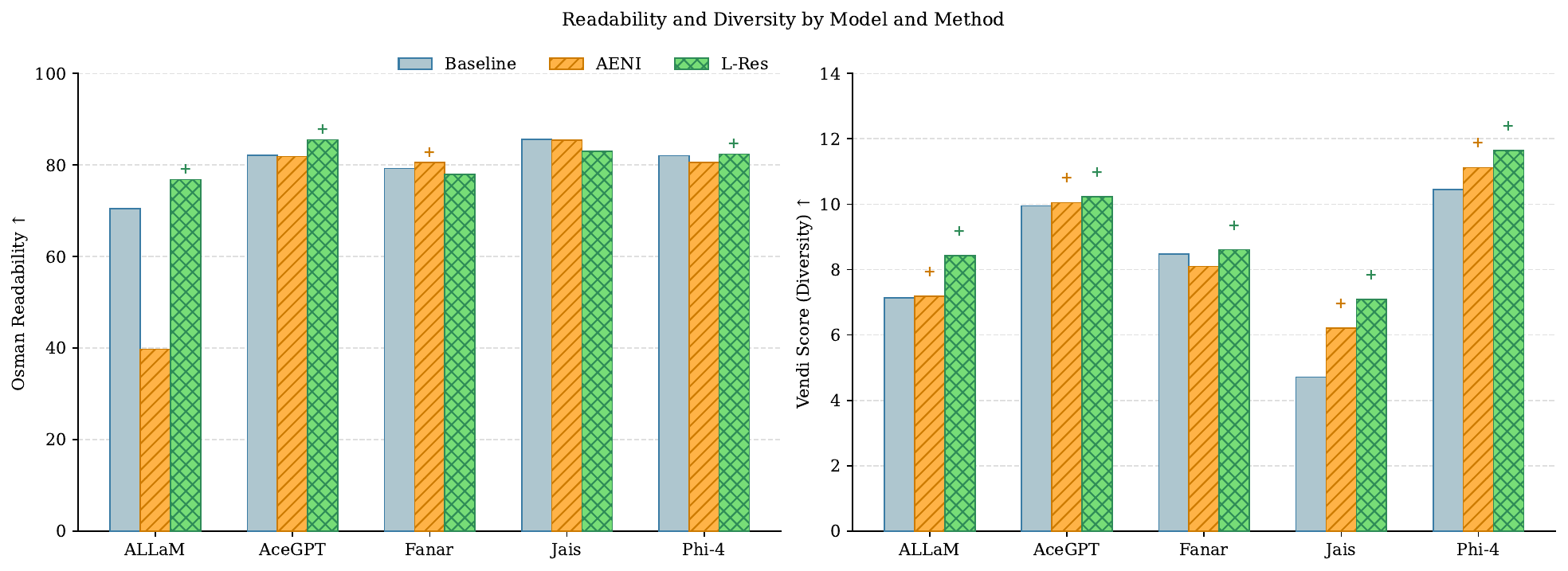}
    \caption{Osman readability score (left) and Vendi Score (right) for Baseline, AENI, and L-Res across all five models. A + marker above a bar indicates the method exceeds the Baseline for that model. Error bars are omitted for clarity.}
    \label{fig:bar}
\end{figure*}

While Figure \ref{fig:bar} contains a figure comparing Osman Readability and Vendi Scores between Baseline, AENI and L-Res methods.

\section{Complete Diversity Scores}

Table \ref{tab:diversity_split} contains the complete vendi and lexical diversity scores of methods that resulted in a Total modal collapse rate of less than 40\%

\begin{table*}[t]
\centering
\small
\setlength{\tabcolsep}{4pt}
\renewcommand{\arraystretch}{1.1}

\setlength{\tabcolsep}{3pt}
\setlength{\columnsep}{6pt}

\begin{subtable}[t]{0.49\textwidth}
\centering
\begin{tabular}{l l c c}
\toprule
\textbf{Model} & \textbf{Method} & VS & LD \\
\midrule
\multirow{7}{*}{ALLaM 7B}
  & Baseline & 7.14 & 0.748 $\pm$ 0.071 \\
  & HiTemp-$k$ & 7.61 & \textbf{0.949} $\pm$ 0.018 \\
  & HiTemp-$p^\dagger$ & 9.15 & 0.956 $\pm$ 0.013 \\
  \cmidrule(l){2-4}
  & Embed$^\dagger$ & 6.92 & 0.838 $\pm$ 0.097 \\
  & Attn$^\dagger$ & 9.51 & 0.906 $\pm$ 0.036 \\
  & AENI & 7.19 & 0.752 $\pm$ 0.119 \\
  & L-Res & \textbf{8.42} & 0.833 $\pm$ 0.066 \\
\bottomrule
\end{tabular}
\caption{ALLaM 7B}
\end{subtable}
\begin{subtable}[t]{0.49\textwidth}
\centering
\begin{tabular}{l l c c}
\toprule
\textbf{Model} & \textbf{Method} & VS & LD \\
\midrule
\multirow{5}{*}{AceGPT 8B}
  & Baseline$^\dagger$ & 9.95 & 0.942 $\pm$ 0.021 \\
  \cmidrule(l){2-4}
  & Embed & \textbf{10.24} & 0.943 $\pm$ 0.026 \\
  & Attn & 10.21 & 0.940 $\pm$ 0.019 \\
  & AENI & 10.05 & \textbf{0.944} $\pm$ 0.022 \\
  & L-Res & 10.23 & 0.941 $\pm$ 0.022 \\
\midrule
\multirow{3}{*}{Phi-4-mini}
  & Baseline & 10.45 & 0.949 $\pm$ 0.013 \\
  \cmidrule(l){2-4}
  & AENI & 11.12 & \textbf{0.954} $\pm$ 0.012 \\
  & L-Res & \textbf{11.64} & 0.951 $\pm$ 0.017 \\
\bottomrule
\end{tabular}
\caption{AceGPT 8B + Phi-4-mini}
\end{subtable}

\vspace{0.8em}

\begin{subtable}[t]{0.49\textwidth}
\centering
\begin{tabular}{l l c c}
\toprule
\textbf{Model} & \textbf{Method} & VS & LD \\
\midrule
\multirow{7}{*}{Fanar 9B}
  & Baseline & 8.47 & 0.939 $\pm$ 0.018 \\
  & HiTemp-$k^\dagger$ & 9.69 & 0.979 $\pm$ 0.005 \\
  & HiTemp-$p$ & \textbf{9.40} & \textbf{0.979} $\pm$ 0.005 \\
  \cmidrule(l){2-4}
  & Embed$^\dagger$ & 8.92 & 0.976 $\pm$ 0.012 \\
  & Attn$^\dagger$ & 8.64 & 0.953 $\pm$ 0.013 \\
  & AENI$^\dagger$ & 8.10 & 0.950 $\pm$ 0.019 \\
  & L-Res & 8.61 & 0.940 $\pm$ 0.024 \\
\bottomrule
\end{tabular}
\caption{Fanar 9B}
\end{subtable}
\hspace{0.01\textwidth}
\begin{subtable}[t]{0.49\textwidth}
\centering
\begin{tabular}{l l c c}
\toprule
\textbf{Model} & \textbf{Method} & VS & LD \\
\midrule
\multirow{7}{*}{Jais 8B}
  & Baseline & 4.71 & 0.762 $\pm$ 0.139 \\
  & HiTemp-$k$ & 6.37 & 0.876 $\pm$ 0.067 \\
  & HiTemp-$p$ & 5.87 & 0.848 $\pm$ 0.076 \\
  \cmidrule(l){2-4}
  & Embed$^\dagger$ & 6.52 & 0.916 $\pm$ 0.052 \\
  & Attn$^\dagger$ & 7.44 & 0.922 $\pm$ 0.031 \\
  & AENI & 6.21 & 0.867 $\pm$ 0.056 \\
  & L-Res & \textbf{7.09} & \textbf{0.928} $\pm$ 0.027 \\
\bottomrule
\end{tabular}
\caption{Jais 8B}
\end{subtable}

\vspace{0.8em}

\vspace{0.5em}
\caption{%
Diversity metrics for all conditions with ${<}\,40\%$ modal collapse.
\textbf{VS}: Vendi Score.
\textbf{LD}: Lexical Diversity ($1 - \text{Self-BLEU}$).
\textbf{Bold}: best per model under ${\leq}\,1$ collapse.
\textsuperscript{$\dagger$}: ${\geq}\,1$ collapse.}
\label{tab:diversity_split}

\end{table*}

\section{Quality Scores}

Table \ref{table:quality} contains the complete quality scores for each run with the exception of Phi-4-mini Embedding Noise as it failed to produce any coherent output.

\begin{table}[h]
\centering
\begin{tabular}{lccccc}
\toprule
Model & Baseline & HiTemp-k & HiTemp-p & Embed & Attn \\
\midrule
ALLaM  
& $7.23 \pm 0.81$ 
& $4.89 \pm 1.80$ 
& $5.13 \pm 1.97$ 
& $4.09 \pm 2.89$ 
& $2.55 \pm 1.24$ \\

AceGPT 
& $5.78 \pm 1.61$ 
& $0.42 \pm 1.03$
& $0.60 \pm 1.24$
& $4.31 \pm 1.50$ 
& $4.88 \pm 1.22$ \\

Fanar  
& $7.97 \pm 0.66$ 
& $4.32 \pm 2.00$ 
& $3.67 \pm 1.11$ 
& $3.86 \pm 2.74$ 
& $4.77 \pm 2.76$ \\

Jais   
& $8.08 \pm 0.92$ 
& $8.18 \pm 0.74$ 
& $7.95 \pm 0.75$ 
& $6.18 \pm 1.80$ 
& $6.58 \pm 1.85$ \\

Phi-4-mini
& $4.64 \pm 1.18$ 
& $2.59 \pm 1.51$
& $1.99 \pm 0.70$
& -- 
& $2.41 \pm 1.02$ \\
\midrule
Model & AENI & L-Res \\
\midrule
ALLaM  
& $6.73 \pm 1.05$ 
& $7.30 \pm 0.80$ \\

AceGPT 
& $5.29 \pm 1.38$ 
& $5.11 \pm 1.47$ \\

Fanar  
& $6.39 \pm 1.38$ 
& $8.33 \pm 0.71$ \\

Jais   
& $7.88 \pm 0.96$ 
& $6.97 \pm 1.36$ \\

Phi-4-mini
& $4.35 \pm 1.29$ 
& $3.67 \pm 1.15$ \\
\bottomrule
\end{tabular}
\caption{Quality mean $\pm$ standard deviation for each model and method}
\label{table:quality}
\end{table}

\end{document}